\documentclass[10pt,twocolumn,letterpaper]{article}

\usepackage{iccv}              

\definecolor{iccvblue}{rgb}{0.21,0.49,0.74}
\usepackage[pagebackref,breaklinks,colorlinks,allcolors=iccvblue]{hyperref}
\usepackage[accsupp]{axessibility}  
\usepackage{amsthm}

\usepackage{amsmath}
\usepackage{algorithm}
\usepackage{algorithmic}
\usepackage{multirow}
\newtheorem{lemma}{Lemma}[section]  

\def\paperID{8047} 
\def\confName{ICCV}
\def\confYear{2025}

\title{Balanced Sharpness-Aware Minimization for Imbalanced Regression}

\author{Yahao Liu\textsuperscript{1} \quad \space\space\space\space Qin Wang\textsuperscript{2} \quad \space\space\space\space    Lixin Duan\textsuperscript{1} \quad \space\space\space\space  Wen Li\textsuperscript{1}\thanks{The corresponding author} \space\space\space\space  \\
       $^1$ University of Electronic Science and Technology of China \quad $^2$ ETH Z\"urich \\
      {\tt\small \{lyhaolive, wangqin.ee, lxduan, liwenbnu\}@gmail.com}
   }

\begin{document}
\maketitle
\begin{abstract}
Regression is fundamental in computer vision and is widely used in various tasks including age estimation, depth estimation, target localization, \etc However, real-world data often exhibits imbalanced distribution, making regression models perform poorly especially for target values with rare observations~(known as the imbalanced regression problem). In this paper, we reframe imbalanced regression as an imbalanced generalization problem. To tackle that, we look into the loss sharpness property for measuring the generalization ability of regression models in the observation space. Namely, given a certain perturbation on the model parameters, we check how model performance changes according to the loss values of different target observations. We propose a simple yet effective approach called Balanced Sharpness-Aware Minimization~(BSAM) to enforce the uniform generalization ability of regression models for the entire observation space. In particular, we start from the traditional sharpness-aware minimization and then introduce a novel targeted reweighting strategy to homogenize the generalization ability across the observation space, which guarantees a theoretical generalization bound. Extensive experiments on multiple vision regression tasks, including age and depth estimation, demonstrate that our BSAM method consistently outperforms existing approaches. The code is available \href{https://github.com/manmanjun/BSAM_for_Imbalanced_Regression}{here}.

\end{abstract}    
\section{Introduction}
\label{sec:intro}

Regression tasks are fundamental in computer vision~\cite{yi2014age,tompson2014joint,lee2019image,li2021human}, encompassing various applications from age estimation to depth estimation. Unlike classification which predicts discrete categories, regression tasks require models to learn more precise continuous mappings, inherently more challenging to optimize and generalize. This challenge is further compounded in real-world scenarios where data imbalance is prevalent. For instance, in age estimation tasks~\cite{moschoglou2017agedb, rothe2015dex}, data from elderly subjects is typically more scarce compared to middle-aged and young subjects. Similar patterns of imbalance persist across various vision tasks, from depth estimation~\cite{silberman2012indoor} where certain depth ranges dominate the data distribution, to image quality assessment~\cite{richhf} where extreme quality scores are rarely observed. 

\begin{figure}[t]
\centering
   \includegraphics[width=1.0\linewidth]{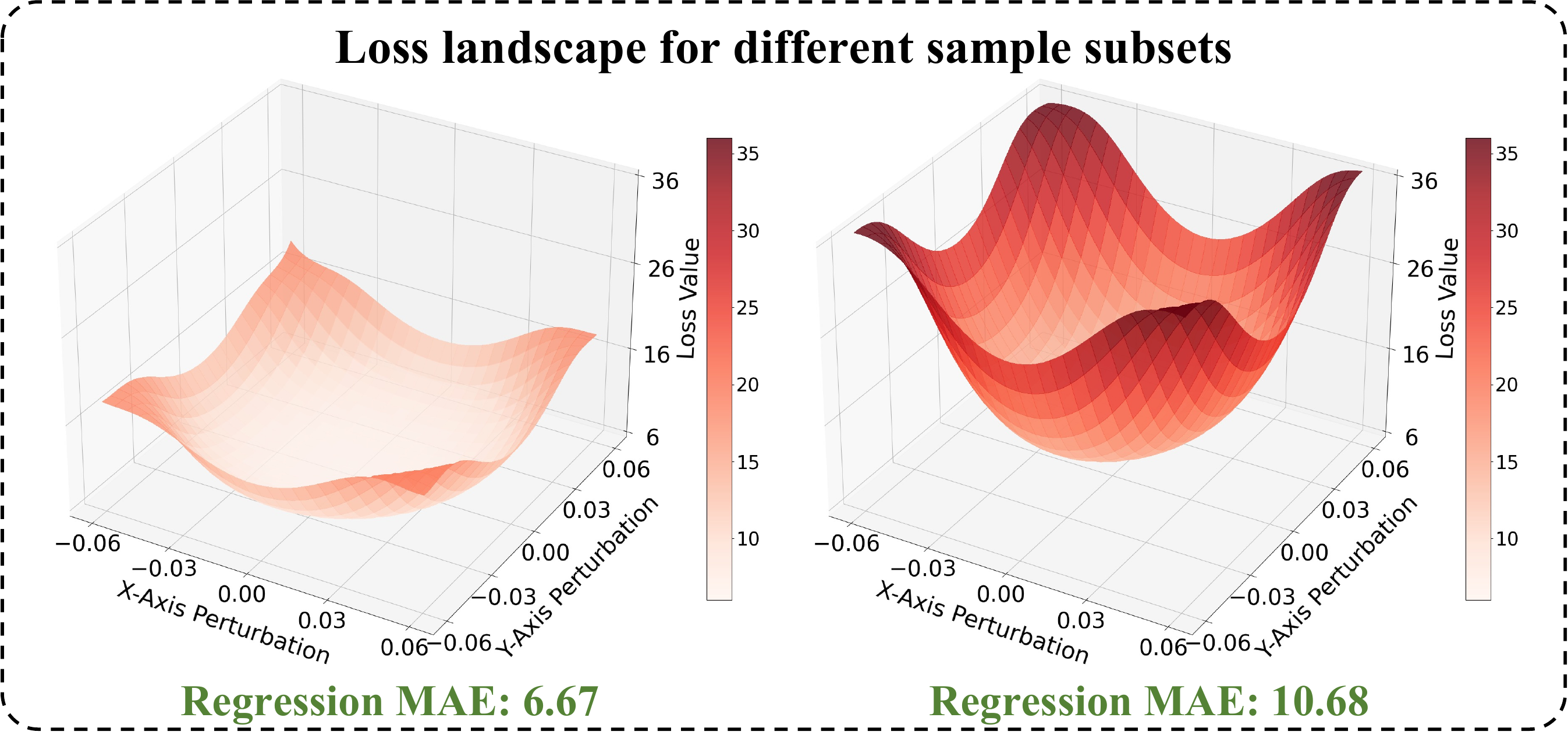}
   \caption{Visualization of loss landscapes for different sample subsets in AgeDB-DIR dataset. \textbf{Left}: Loss landscape computed over the entire dataset, exhibiting relatively smooth geometry with a lower MAE of $6.67$, indicating better generalization. \textbf{Right}: Loss landscape for samples from low-density regions shows significantly increased sharpness and higher sensitivity to parameter perturbations, resulting in degraded generalization performance~(MAE: $10.68$). This stark contrast in both landscape geometry and quantitative metrics demonstrates the inherent challenge of maintaining consistent generalization ability across different density regions in imbalanced regression.}
\label{fig:loss_landscape}
\vspace{-5mm}
\end{figure} 

Prior research has established that regression tasks necessitate careful consideration of label continuity and inter-label relationships~\cite{yang2021delving, zha2023rank, keramati2024conr}. Various strategies have been proposed to address this challenge:  \cite{yang2021delving} propose using Gaussian kernel smoothing during sample reweighting to maintain continuity between label weights. Others emphasize the importance of feature continuity during model optimization, employing contrastive learning~\cite{zha2023rank, keramati2024conr} and rank-based~\cite{zha2023rank} approaches to impose additional constraints on model features. 

While existing methods have shown promise, a fundamental challenge persists: the degraded model generalization under imbalanced data distributions, where test performance consistently deteriorates for low-density values that are underrepresented during training. To analyze this phenomenon, we visualize the loss landscape of imbalanced regression models following the visualization methodology proposed in~\cite{visualloss}, as illustrated in Figure~\ref{fig:loss_landscape}. Namely, by applying a controlled perturbation on the parameters of a regression model trained from imbalanced data, we can visualize the resulting loss variation of the test observations. When focusing specifically on samples from low-density regions of the data distribution, we observe significantly degraded performance in MAE, indicating poor model generalization in these regions, along with markedly increased sharpness in loss landscapes compared to the overall dataset. This observation aligns with established studies~\cite{keskar2017on, foret2021sharpnessaware} that demonstrate strong connections between loss landscape geometry and model generalization, where flatter minima often correlate with superior generalization performance.

Motivated by these findings, we propose Balanced Sharpness-Aware Minimization~(BSAM) for imbalanced regression, a novel optimization framework that addresses the challenge of imbalanced regression through the lens of loss landscape geometry.  While traditional Sharpness-Aware Minimization~(SAM)~\cite{foret2021sharpnessaware} has demonstrated significant improvements in model generalization by seeking flat minima of the loss landscape, our analysis reveals that its direct application to imbalanced regression tasks is susceptible to distributional biases. Specifically, SAM's uniform treatment of all samples in the perturbation step leads to optimization trajectories biased toward high-density regions of the target distribution. 

To address this, BSAM extends the theoretical foundations of SAM in two crucial aspects. First, rather than simply pursuing flat minima, BSAM aims to achieve consistent flatness of loss landscape across the entire observation space. Second, we introduce a targeted reweighting mechanism that dynamically adjusts each sample's influence during the perturbation calculation process. This approach effectively balances the contribution of samples across the entire target distribution spectrum, preventing the dominance of frequently observed target values while simultaneously ensuring robust generalization across all regions. Our framework maintains algorithmic simplicity while avoiding the complexity often associated with specialized loss functions or distribution smoothing techniques.
Through extensive experimentation across multiple vision regression tasks, including age estimation and depth prediction, we demonstrate that BSAM consistently achieves state-of-the-art performance. Our empirical results validate the effectiveness of combining sharpness-aware optimization with targeted reweighting strategies.
The main contributions of our work can be summarized as follows:

\begin{itemize}
\item We reframe the imbalanced regression problem from a novel perspective of generalization ability, revealing the connection between loss landscape geometry and model performance of the target distribution, especially in low-density regions.
\item We propose Balanced Sharpness-Aware Minimization~(BSAM), a simple yet effective framework that addresses the limitations of conventional SAM in imbalanced regression by integrating targeted reweighting mechanisms, derived from our generalization analysis.
\item We validate the effectiveness of BSAM through comprehensive experiments including age estimation and depth prediction, achieving superior performance across multiple vision regression tasks.
\end{itemize}

\section{Related Work}
\label{sec:Related Work}
\subsection{Regression for Vision Tasks}

Regression tasks are fundamental in computer vision, underpinning a wide range of applications, including age estimation~\cite{yi2014age,rothe2018deep, moschoglou2017agedb}, depth estimation~\cite{silberman2012indoor,eigen2014depth}, pose estimation~\cite{martinez2017simple, li2021human}, and image quality assessment~\cite{lee2019image,richhf}. Traditional regression methods typically employ $l_1$, $l_2$, and Huber loss~\cite{huber1992robust} to learn a continuous mapping between input images and target values. Recent research has focused on the regression-specific characteristics such as the ordinal relationships inherent to the dataset have been utilized to design more effective loss functions and decompose the regression task into multiple binary classification tasks~\cite{rothe2015dex, niu2016ordinal, fu2018deep,shah2022label}. Additionally, in the field of pose estimation, \cite{tompson2014joint, bulat2016human} reformulate the regression problem as a segmentation task by generating heatmaps to represent continuous variables spatially. However, these approaches typically assume balanced data distributions, which rarely hold in real-world scenarios.

\subsection{Imbalanced Regression}
Imbalanced regression has received comparatively less attention than its classification counterpart. However, many real-world vision tasks such as depth estimation~\cite{silberman2012indoor}, age estimation~\cite{rothe2018deep, moschoglou2017agedb}, and image quality assessment~\cite{richhf} often exhibit long-tail distributions, resulting in a severe imbalance where certain target values are significantly underrepresented. Early methods largely relied on SMOTE-based algorithms~\cite{chawla2002smote}, which use linear interpolation for data augmentation. Recently, significant progress has been made in tackling this challenge. \cite{yang2021delving} proposed a comprehensive benchmark for imbalanced regression and introduced label and feature smoothing techniques based on local similarities. The connection between regression and classification losses has been investigated in several studies~\cite{zhang2023improving, pintea2023step,xiong2024deep}; for instance, \cite{zhang2023improving} demonstrated that enforcing constraints to encourage high-entropy feature spaces can enhance regression performance. Statistical approaches have also emerged in this area~\cite{ren2022balanced, wang2024variational}. For example, \cite{ren2022balanced} modeled regression predictions as a Gaussian distribution to design a balanced mean squared error~(MSE) loss, addressing the imbalance in a probabilistic framework. Additionally, feature-label consistency constraints for contrastive learning~\cite{zha2023rank, keramati2024conr} and rank-based constraints~\cite{gong2022ranksim} have shown promising results in this domain. Among these, the Rank-N-contrast method~\cite{zha2023rank} stands out, achieving state-of-the-art performance through a two-stage training process: initially training a feature extractor using advanced contrastive learning techniques, followed by a separate regressor training phase. Despite these advances, we reframe imbalanced regression as a generalization problem, addressing the inconsistent model behavior between training and test distributions across different value ranges.

\subsection{Loss Landscape}
The geometry of the loss landscape plays a crucial role in understanding the generalization capabilities of deep neural networks. Extensive research has demonstrated that flatter minima typically correlate with superior generalization performance~\cite{keskar2017on, chaudhari2019entropy, jiang2019fantastic}.  Leveraging this insight, Sharpness-Aware Minimization~(SAM)~\cite{foret2021sharpnessaware} has been introduced to explicitly encourage flatness by minimizing the worst-case perturbation in the loss landscape. SAM has demonstrated significant improvements in robustness and generalization across various tasks~\cite{cha2021swad,rangwani2022closer,bahri2021sharpness}. Recent adaptations of SAM have extended its application to imbalanced learning scenarios. Notably, ImbSAM~\cite{zhou2023imbsam} and CC-SAM~\cite{zhou2023class} have tailored the approach for class-imbalanced datasets. However,  these methods primarily address classification challenges, leaving the challenging domain of imbalanced regression relatively unexplored (detailed analysis in Section~\ref{sec:discussion}).

\section{Methodology}
\label{Methodology}

In imbalanced regression tasks, we are provided with a training set $\mathcal{S}_{tr} = \{(\mathbf{x}_i, y_i)\}_{i=1}^N$ drawn from a distribution $\mathcal{D}_{tr}$, where $\mathbf{x}_i \in \mathbb{R}^{H \times W \times C} $ represents an RGB image of height $H$, width $W$, and channels $C$, and $y_i \in \mathbb{R}$ denotes the corresponding continuous label. Unlike classification tasks, the label space $\mathcal{Y}$ in regression is continuous, bounded by a lower bound $L$ and an upper bound $U$, such that: $\mathcal{Y} = \{y \mid L \leq y \leq U\}$, where \(y\) is the continuous target value associated with each input data point. In general, a regression network can be described as a function $f_{\theta}: \mathbb{R}^{H \times W \times C} \rightarrow \mathbb{R}$, which maps the input $\mathbf{x}$ to a continuous output $y$. Here, $\theta$ represents the model parameters, which are learned by minimizing the loss function $\mathcal{L}_{tr}(\theta) = \frac{1}{N} \sum_{i=1}^N \ell(f_{\theta}(\mathbf{x}_i), y_i)$ over the training set. The common choices for $\ell$ are $l_1$ or $l_2$. To quantify the imbalance in regression tasks, we use a histogram over the label space \(\mathcal{Y}\) to represent the distribution of target values. The histogram divides the continuous label range from \(L\) to \(U\) into a series of intervals $b_1,...,b_K$, each accumulating the frequency of samples falling within that interval. 

This systematic partitioning enables us to quantitatively assess the distribution of samples across the continuous label space. Imbalance occurs when substantial variation in sample density exists across different intervals, manifesting as regions that are either sparsely populated~(rare values) or densely populated~(common values) with samples.
The primary learning objective of imbalanced regression is to develop a model that can accurately predict values across the entire range of the target variable, ensuring low loss on $\mathcal{L}_{test}(\theta)$  balanced data distribution $\mathcal{D}_{test}$.

\subsection{Generalization Analysis for Imbalanced Regression}

A regression model trained with imbalanced data can suffer from deteriorated generalization ability. As shown in Figure~\ref{fig:loss_landscape}, the test performance consistently deteriorates for low-density values that are underrepresented during training. To further quantify this generalization ability difference, we analyze the loss landscape for the overall dataset and the low-density region. Given a controlled perturbation on the model parameter, the loss change is much more rapid for the low-density samples~(right) than the dataset average~(left). All of these demonstrate the inherent challenge of maintaining consistent generalization ability across different density regions in imbalanced regression.

To address these generalization challenges, we adopt the framework of Sharpness-Aware Minimization~\cite{foret2021sharpnessaware} which improves model generalization by seeking parameter values whose entire neighborhoods have uniformly low training loss. The theoretical foundation for this approach is established through a theorem that bounds generalization ability based on neighborhood-wise training loss:
\begin{lemma} \label{key_theory}
For any $\rho > 0$, with high probability over training set $\mathcal{S}$ generated from distribution $\mathcal{D}$,
\begin{align}
\mathcal{L}_\mathcal{D}(\theta) &\leq \max_{\|\epsilon\|_2 \leq \rho} \mathcal{L}_\mathcal{S}(\theta + \epsilon) + h(\|\theta\|_2^2/\rho^2),
\end{align}
where $h: \mathbb{R}_+ \rightarrow \mathbb{R}_+ $ is a strictly increasing function.
\end{lemma}
Examining the right side of the inequality reveals two key components. The first term describes the worst-case loss within a $\rho$-radius neighborhood of the current parameters $\theta$, effectively characterizing the local sharpness of the loss landscape. The second term $h(\|\theta\|_2^2 / \rho^2)$ represents a regularization component typically controlled through weight decay in deep learning practice. This theorem establishes a fundamental relationship: when the training set $\mathcal{S}$ is sampled from distribution $\mathcal{D}$, the generalization error on distribution $\mathcal{D}$ can be bounded by the maximum perturbation error within a neighborhood on the training set $\mathcal{S}$. 

However, this formulation reveals a critical limitation in imbalanced regression settings. According to Lemma~\ref{key_theory}, when measuring model sharpness using the training set $\mathcal{S}_{tr}$, SAM can only guarantee generalization to the training distribution $\mathcal{D}_{tr}$. 
This limitation becomes particularly acute in imbalanced scenarios, where our target is the balanced test distribution $\mathcal{D}_{te}$. Under such distribution shifts~($\mathcal{D}_{tr} \neq \mathcal{D}_{te}$), conventional SAM's generalization guarantees deteriorate significantly, especially for underrepresented samples. The inadequate exploration of the loss landscape in these underrepresented regions leads to unreliable sharpness estimates and compromised generalization performance. These theoretical insights highlight the necessity for a more sophisticated approach that explicitly addresses the distribution shift between training and testing environments.

\subsection{Balanced Sharpness-Aware Minimization}

As analyzed in the previous section, the conventional SAM approach, while effective for standard learning scenarios, fails to provide adequate generalization guarantees for the balanced test set when faced with the imbalance of the training set. To address this, we propose Balanced Sharpness-Aware Minimization~(BSAM), a novel approach that explicitly accounts for the target balanced distribution in its optimization framework.

To establish our method, we provide an analytical solution for computing the maximum perturbation $\epsilon^*$ in Lemma~\ref{key_theory}. Specifically, given a parameter vector $\theta$
and a small perturbation radius $\rho$, the worst-case perturbation within the $\rho$-ball can be approximated through first-order Taylor expansion of $\mathcal{L}_\mathcal{S}(\theta+\epsilon)$: 
\begin{align}
\epsilon^* &= \underset{\|{\epsilon}\|_p \leq \rho}{\operatorname{argmax}}\,  \mathcal{L}_\mathcal{S}(\theta + \epsilon) \\
&\approx \underset{\|{\epsilon}\|_p \leq \rho}{\operatorname{argmax}}\,  [\mathcal{L}_\mathcal{S}(\theta) + \epsilon^T \nabla_\theta \mathcal{L}_\mathcal{S}(\theta)] \\
& =\rho\cdot \operatorname{sign}\left(\nabla_\theta \mathcal{L}_{\mathcal{S}}\left(\theta\right)\right) \cdot\frac{\left|\nabla_\theta \mathcal{L}_{\mathcal{S}}\left(\theta\right)\right|^{q-1}}{\left\|\nabla_\theta \mathcal{L}_{\mathcal{S}}\left(\theta\right)\right\|_q^{q / p}},
\end{align}

where $\rho \geq 0$ is a hyperparameter and $1/p + 1/q = 1$. The last equation gives the closed-form solution for the maximum perturbation $\epsilon^*$ that maximizes the local loss increase by the Dual Norm Problem solution. Once this worst-case perturbation $\epsilon^*$ is determined, we can update the model parameters $\theta_t $ of step $t$ using the following optimization step:
\begin{equation}
\theta_{t+1} = \theta_t - \alpha_t \cdot \nabla \mathcal{L}_\mathcal{S}(\theta_t + \epsilon^*) ,
\end{equation}
where $\alpha_t$ is the learning rate at step $t$. Here, we have omitted the explicit weight decay term for simplicity. 

Notably, the primary impact of distribution discrepancy manifests in the computation of the maximum perturbation $\epsilon^*$. Therefore, to effectively address the imbalance issue and ensure generalization to the balanced test distribution $\mathcal{D}_{te}$, we propose a simple yet effective reweighting strategy that incorporates balance-aware weights directly into the perturbation computation process.

Specifically, to bridge the gap between training and testing label distributions, we first examine the relationship between Empirical Risk Minimization~(ERM) $\mathcal{L}_{erm}$\footnote{
$\mathcal{L}_{erm} = \mathbb{E}_{(\mathbf{x},y) \sim \mathcal{S}_{tr}} \ell\left(f_\theta\left(\mathbf{x}\right), y\right)$}
and Balanced Error~(BE) $\mathcal{L}_{be}$. Let $P_{tr}(k)$ denote the empirical distribution of interval $k$ in the training set and $P_{te}(k)$ represents the desired uniform distribution for our test set. Through importance reweighting, we can bridge these two objectives:
\begin{align}
\mathcal{L}_{be}(\theta) &= \sum_{k=1}^K P_{te}(k) \cdot\mathbb{E}_{(\mathbf{x},y) \sim S^{b_k}}\ell\left(f_\theta\left(\mathbf{x}\right), y\right) \\
&= \sum_{k=1}^K P_{te}(k) \cdot\frac{P_{tr}(k)}{P_{tr}(k)} \cdot\mathbb{E}_{(\mathbf{x},y) \sim S^{b_k}}\ell\left(f_\theta\left(\mathbf{x}\right), y\right) \\
&= \mathbb{E}_{(\mathbf{x},y) \sim \mathcal{S}_{tr}} w(k(y)) \cdot\ell\left(f_\theta\left(\mathbf{x}\right), y\right),
\end{align}
where $w(k(y)) = \frac{P_{te}(k(y))}{P_{tr}(k(y))}$ represents the importance weight for samples from bin $k$. Building upon this insight, we incorporate this reweighting mechanism into the computation of maximum perturbation $\epsilon$:
\begin{align}
\boldsymbol{\epsilon} & =\underset{\|\boldsymbol{\epsilon}\|_p \leq \rho}{\operatorname{argmax}} \, \mathbb{E}_{(\mathbf{x}, y) \sim \mathcal{S}_{tr}}\left[w(k(y))\cdot\ell \left(f_{\theta+\epsilon}(\mathbf{x}), y\right)\right] \\
& \approx \rho \cdot \operatorname{sign} (\nabla \mathcal{L}^w_{\mathcal{S}_{tr}}(\theta)) \cdot\frac{\left|\nabla \mathcal{L}^w_{\mathcal{S}_{tr}}(\theta)\right|^{q-1}}{\left\|\nabla \mathcal{L}^w_{\mathcal{S}_{tr}}(\theta))\right\|_q^{q / p}}, \label{eq:weight_perturbation}
\end{align}
\begin{equation}
\label{eq:weight_loss}
\mathcal{L}^w_{\mathcal{S}_{tr}}(\theta)=\frac{1}{N} \sum_{i=1}^N w\left(k(y_i)\right)\cdot \ell\left(f_\theta\left(\mathbf{x}_i\right), y_i\right).
\end{equation}

It is important to note that BSAM differs fundamentally from traditional loss reweighting approaches. While loss reweighting modifies the optimization objective used in parameter updates, BSAM introduces the targeted reweighting specifically in the perturbation computation step, making it independent of the specific form of the loss function $\mathcal{L}$ used in parameter updates, which we will verify in the experimental section.

\begin{algorithm}[t]
\caption{Balanced Sharpness-Aware Minimization (BSAM)}
\label{alg:bsam}
\begin{algorithmic}[1]
\REQUIRE Training set $\mathcal{S}_{tr}$, initial parameters $\theta_0$, learning rate $\{\alpha_t\}_{t=1}^T$, neighborhood size $\rho$, number of bins $K$
\STATE Divide label space into $K$ equal-width interval $b_1,...,b_K$
\STATE Calculate bin frequencies $\{N_k\}_{k=1}^K$ from training set
\STATE Compute importance weights $w(k)$ for each bin $k$
\FOR{$t = 1$ to $T$}
    \STATE Sample a mini-batch $\mathcal{B}_t$ from $\mathcal{S}_{tr}$
    \STATE Compute weighted loss function by Eq.~\ref{eq:weight_loss} 
    \STATE Calculate the perturbation $\epsilon^*$ by Eq.~\ref{eq:weight_perturbation}
    \STATE Update parameters:
    \STATE $\theta_{t+1} = \theta_t - \alpha_t\cdot\nabla \mathcal{L}(\theta_t + \epsilon^*)$
\ENDFOR
\RETURN Final model parameters $\theta_T$
\end{algorithmic}
\end{algorithm}
\subsection{Overall algorithm and discussion} \label{sec:discussion}

Our analysis reveals that there is an inherent disparity in loss landscape geometry across different density regions, leading to inconsistent generalization ability across the target distribution. This observation motivates us to develop an optimization framework that maintains uniform generalization ability across the entire observation space by seeking loss sharpness with consistent flatness regardless of the training set distribution density. Based on the analysis above, we present the complete algorithm of Balanced Sharpness-Aware Minimization (BSAM) for imbalanced regression tasks. The overall procedure is summarized in Algorithm~\ref{alg:bsam}.

\paragraph{Discussion.} Our BSAM differs from existing SAM variants~\cite{zhou2023imbsam,zhou2023class} for imbalanced learning in several key aspects. First, while both \cite{zhou2023imbsam} and \cite{zhou2023class} focus on classification tasks, BSAM is specifically designed for regression problems. Second, \cite{zhou2023imbsam} selectively computes perturbations using only minority classes based on a predefined threshold. However, as indicated by Lemma~\ref{key_theory}, this approach potentially compromises the generalization capability across the entire data distribution, which we empirically verify in our experiments. On the other hand, \cite{zhou2023class} calculates perturbations separately for each class, which cannot guarantee that the perturbation maximizes the loss concerning the entire target distribution $\mathcal{D}_{te}$. Moreover, its requirement to iterate through all classes during each update becomes computationally intractable for regression tasks where the label space is continuous. In contrast, BSAM employs a simple yet effective targeted weighted perturbation strategy that doesn't introduce additional computational complexity to the original SAM while effectively capturing the local geometry of the loss landscape for $\mathcal{D}_{test}$.

\section{Experiments}
\label{Experiments}

\begin{table*}[t]
\footnotesize
\centering
\caption{\textbf{Main results on \textit{AgeDB-DIR} benchmark.} Results marked with $\star$ are directly quoted from their original paper while results marked with $\dagger$ are obtained through our reproduction and RNC~\cite{zha2023rank} following the RNC training protocol.}
\label{table:agedb}
\begin{tabular}{@{}lllllcllll@{}}
\specialrule{2\heavyrulewidth}{\abovetopsep}{\belowbottomsep}
& \multicolumn{4}{c}{\textbf{MAE $\downarrow$}} & \phantom{abc}& \multicolumn{4}{c}{\textbf{GM $\downarrow$}} \\
\cmidrule{2-5} \cmidrule{7-10} 
& All & Man. & Med. & Few &&  All & Man. & Med. & Few \\
\midrule[1.2pt]
LDS~\cite{yang2021delving}$^\star$&7.42&6.83&8.21&10.79&&4.85&4.39&5.80&7.03\\
FDS~\cite{yang2021delving}$^\star$&7.55&6.99&8.40&10.48&&4.82&4.49&5.47&6.58\\
RankSim~\cite{gong2022ranksim}$^\star$&6.91&6.34&7.79&9.89&&4.28&3.92&4.88&6.89\\
ConR~\cite{keramati2024conr}+LDS~\cite{yang2021delving}$^\star$&7.16&6.61&7.97&9.62&&4.51&4.21&4.92&5.87\\
ConR~\cite{keramati2024conr}+FDS~\cite{yang2021delving}$^\star$&7.08&6.46&7.89&9.80&&4.31&4.01&5.25&6.92\\
\midrule[1.2pt]
RankSim~\cite{gong2022ranksim}$^\dagger$ &6.51  & - & - & - && - & - & - & -  \\ 
RNC~\cite{zha2023rank}$^\dagger$ &6.14 & - & - & -  && - & - & - & -  \\
LDS~\cite{yang2021delving}$^\dagger$&6.350  & 5.925  &7.078 & 8.355 && 3.963 & 3.721 & 4.441 & \textbf{5.249}  \\
Ordinal Entropy~\cite{zhang2023improving}$^\dagger$&6.360&5.778  & 7.059 &9.921  && 3.987 &\textbf{3.656} & 4.382& 6.958  \\
\midrule[1.2pt]
Vanilla$^\dagger$ &6.690  &5.959  & 7.740   &10.688  && 4.254 & 3.734 &5.281  &8.021  \\
SQINV$^\dagger$ & 6.391 & 5.955 &  7.155 & 8.390 && 4.039 & 3.774 & 4.577 &5.425 \\
BSAM & \textbf{6.067}  & \textbf{5.801} & \textbf{6.304}  & \textbf{7.928} &&  \textbf{3.895}& 3.748 & \textbf{3.925} & 5.473\\
\bottomrule[1.5pt]
\end{tabular}
\end{table*}

\begin{table*}[t]
\footnotesize
\centering
\caption{\textbf{Main results on \textit{IMDB-WIKI-DIR} benchmark.} Results marked with $\star$ are directly quoted from their original paper while results marked with $\dagger$ are obtained through our reproduction following the RNC  training protocol.}
\label{table:imdb}
\begin{tabular}{@{}lllllcllll@{}}
\specialrule{2\heavyrulewidth}{\abovetopsep}{\belowbottomsep}
& \multicolumn{4}{c}{\textbf{MAE $\downarrow$}} & \phantom{abc}& \multicolumn{4}{c}{\textbf{GM $\downarrow$}} \\
\cmidrule{2-5} \cmidrule{7-10} 
& All & Man. & Med. & Few &&  All & Man. & Med. & Few \\
\midrule[1.2pt]
LDS~\cite{yang2021delving}$^\star$&7.83&7.31&12.43&22.51&&4.42&4.19&7.00&13.94\\
FDS~\cite{yang2021delving}$^\star$&7.83&7.23&12.60&22.37&&4.42&4.20&6.93&13.48\\
RankSim~\cite{gong2022ranksim}$^\star$&7.42&6.84&12.12&22.13&&4.10&3.87&6.74&12.78\\
ConR~\cite{keramati2024conr}+LDS~\cite{yang2021delving}$^\star$&7.43&6.84&12.38&21.98&&4.06&3.94&6.83&12.89\\
ConR~\cite{keramati2024conr}+FDS~\cite{yang2021delving}$^\star$&7.29&6.90&12.01&21.72&&4.02&3.83&6.71&12.59\\
\midrule[1.2pt]
Ordinal Entropy~\cite{zhang2023improving}$^\dagger$& 7.322 &6.629   & 13.154  & 23.235 && 4.000 & 3.708  & 7.977 & 14.961  \\
RNC~\cite{zha2023rank}$^\dagger$ &7.466  & 6.757  & 13.511  & 23.168 && 4.043  &3.729  & 8.516 &15.540 \\
LDS~\cite{yang2021delving}$^\dagger$&7.214 & 6.686  & 11.491  & 20.659  && 4.030  &3.835  &6.127 &\textbf{12.425} \\
\midrule[1.2pt]
Vanilla$^\dagger$ &7.358  &  6.644 & 13.391   &23.544  &&  4.110 &3.784  &  8.789 &  16.101\\
SQINV$^\dagger$ & 7.040 &  6.508& 11.263   & \textbf{21.301} && 3.921& 3.710 &6.233  &  14.457 \\
BSAM &\textbf{6.811}& \textbf{6.294}  & \textbf{10.823}   & 21.339  && \textbf{3.765} &\textbf{3.580}  & \textbf{5.663} &13.609 \\
\bottomrule[1.5pt]
\end{tabular}
\end{table*}
\subsection{Datasets}
We conduct experiments on three public benchmarks for deep imbalanced regression, including two age estimation datasets and one depth estimation dataset. To ensure fair comparisons, we follow the dataset splits as described in \cite{yang2021delving}. For age estimation, we use bins with an interval of 1 year, while for depth estimation, bins are defined with an interval of 0.1 meters, consistent with previous related works~\cite{yang2021delving,ren2022balanced, gong2022ranksim}.
\begin{itemize}[leftmargin=2em,itemsep=0pt]
    \item \textbf{AgeDB-DIR} is an imbalanced facial age estimation benchmark, derived from the AgeDB dataset~\cite{moschoglou2017agedb}. It comprises $16,488$ manually curated noise-free labeled images, with the training set containing $12,208$ images, and both the validation and test sets containing $2,140$ images each.
    \item \textbf{IMDB-WIKI-DIR} is an imbalanced facial age estimation benchmark, derived from the IMDB-WIKI dataset~\cite{rothe2018deep}. It comprises $213,554$ images semi-automatically collected and annotated from the IMDB and Wikipedia websites. The training set consists of $191,509$ images, while both the validation and test sets contain $11,022$ images each.
    \item \textbf{NYUD2-DIR} is an imbalanced depth estimation benchmark, derived from the NYU Depth Dataset V2~\cite{silberman2012indoor}. It includes $50,688$ images for training and $654$ images for testing. Notably, following the default setting of \cite{yang2021delving}, the test dataset of NYUD2-DIR only considers a randomly selected $9,357$ pixels per bin from the $654$ test images to ensure the test set is balanced, corresponding to the minimum number of pixels in any bin of the test set.
\end{itemize}

\subsection{Implementation Details}
\label{sec:implementation}
For all experiments, unless otherwise specified, we use square-root-inverse weighting for calculating $w(k)$ and set $p=2$ for calculating $\rho$. All experiments are conducted on Tesla V$100$ GPUs with a PyTorch implementation.

\noindent\textbf{AgeDB-DIR and IMDB-WIKI-DIR Benchmark.} Following the experiment setup used in RnC~\cite{zha2023rank}, ResNet-$18$~\cite{he2016deep} is utilized as the backbone, with the same data augmentations applied across all compared methods, including random crop, resize (with random horizontal flip), and color distortions. We use the $l_1$ loss~(Vanilla) combined with the square-root-inverse weighting variant~(SQINV) as the primary optimization objective, maintaining a fixed batch size of $256$. For evaluation, Mean Squared Error~(MAE) and Geometric Mean~(GM) are selected as metrics, quantifying the model’s accuracy and fairness in predictions, respectively. Following the approach in~\cite{ren2022balanced}, we also present the results on the balanced Mean Absolute Error~(bMAE) metric. Following~\cite{yang2021delving}, we divide the target space into three disjoint subsets: many-shot region~(intervals with over $100$ training samples), medium-shot region~(intervals with $20\sim100$ training samples), and few-shot region~(intervals with under $20$ training samples). 

\begin{table*}[t]
\centering
\footnotesize
\caption{\textbf{Main results on \textit{NYUD2-DIR} benchmark.} Results marked with $\star$ are directly quoted from their original paper.}
    \label{table:NYU}
    \begin{tabular}{@{}lccccccccc@{}}
    \specialrule{2\heavyrulewidth}{\abovetopsep}{\belowbottomsep}
    & \multicolumn{4}{c}{\textbf{RMSE$\downarrow$}} & \phantom{abc}& \multicolumn{4}{c}{\textbf{$\delta_1$$\uparrow$}} \\
    \cmidrule{2-5} \cmidrule{7-10} 
    
    & All& Man. & Med. & Few && All& Man. & Med. & Few \\
    \midrule[1.2pt]
        FDS~\cite{yang2021delving}$^\star$& 1.442 &0.615 &0.940 &2.059 &&0.681 &0.760 &0.695 &0.596\\
        LDS~\cite{yang2021delving}$^\star$ &1.387 &0.671 &0.913 &1.954 &&0.672 &0.701& 0.706 &0.630\\
        Balanced MSE~(BNI)~\cite{ren2022balanced}$^\star$ &  1.283 &0.787 &0.870 &1.736 &&0.694 &0.622 &0.806 &0.723\\
        Balanced MSE~(BNI)~\cite{ren2022balanced} + LDS~\cite{yang2021delving}$^\star$&1.319	&0.810	&0.920&	1.820&&	0.681	&0.601	&0.695	&0.648\\
        ConR~\cite{keramati2024conr} + FDS~\cite{yang2021delving}$^\star$ & 1.299 & 0.613 & 0.836 & 1.825 && 0.696 & \textbf{0.800} & 0.819 & 0.701 \\
        ConR~\cite{keramati2024conr} + LDS~\cite{yang2021delving}$^\star$ & 1.323 & 0.786 & \textbf{0.823} & 1.852 && 0.700 & 0.632 & \textbf{0.827} & 0.702 \\
        \midrule
        Vanilla$^\star$ &1.477 &\textbf{{0.591}} &0.952 &2.123 &&0.677 &0.777 &0.693 &0.570       \\ 
        SQINV &1.341	&0.604	&0.832&	1.912&&	0.717	&0.769	&0.752	&0.653 \\
        BSAM & \textbf{{1.272}}	&0.728	&1.046&	\textbf{{1.705}}&&	\textbf{{0.727}}	&0.742	&0.695	&\textbf{{0.724}}\\

        \specialrule{2\heavyrulewidth}{\abovetopsep}{\belowbottomsep}
             
    \end{tabular}  
\end{table*}

\noindent\textbf{NYUD2-DIR Benchmark.} Following \cite{yang2021delving,gong2022ranksim}, we adopt ResNet-$50$~\cite{he2016deep} as the backbone, integrating it within an encoder-decoder architecture~\cite{Hu2019RevisitingSI}. The $l_2$ loss~(Vanilla) with the square-root-inverse weighting variant~(SQINV) is applied as the optimization objective, with a batch size of $32$. Evaluation is conducted using root mean squared error~(RMSE) and threshold accuracy $\delta_1$. 

We show the label distributions for three datasets and the detailed formulations of our evaluation metrics and reweighting strategies for $w$ in the supplementary material.
\subsection{Analysis}

\subsubsection{Comparisons with state-of-the-art methods}

We compare our proposed methods with previous state-of-the-art imbalance regression approaches on three benchmarks in Table~\ref{table:agedb}, Table~\ref{table:imdb}, and Table~\ref{table:NYU} respectively. Specifically, we selected two baseline methods, \textit{Vanilla} and \textit{SQINV}, and multiple state-of-the-art regression learning schemes: 1) distribution rebalance methods include the distribution smoothing~\cite{yang2021delving} methods~(LDS and FDS) and the Balanced MSE~\cite{ren2022balanced}, 2) feature space constraints methods include the contrastive learning~\cite{keramati2024conr,zha2023rank}, rank-based constraints~\cite{gong2022ranksim} and the entropy constraints~\cite{zhang2023improving} methods. 

Given that the recent RnC method~\cite{zha2023rank} established a comprehensive experimental protocol for data augmentations and model training strategies in age prediction tasks, which leads to stronger baselines, we conducted our age prediction experiments based on RnC training protocol as mentioned in Section~\ref{sec:implementation}. For a fair comparison, we report the baseline results quoted from~\cite{zha2023rank}, which explains the missing metrics for some methods. We also re-produce other relevant methods following the identical RnC training strategies. We also present the original results of these methods in Table~\ref{table:agedb} and~\ref{table:imdb} for comprehensive evaluation.

\noindent\textbf{AgeDB-DIR Benchmark.} We evaluate our method on the AgeDB-DIR benchmark and compare it with state-of-the-art approaches. As shown in Table~\ref{table:agedb}, our method achieves superior performance across different metrics. Specifically, our BSAM with SQINV achieves the lowest MAE of $6.067$ on All settings and GM of $3.895$, surpassing previous methods by a clear margin. The improvement is particularly significant in Few-shot scenarios, where our method reduces the MAE from $10.688$ to $7.928$, demonstrating its effectiveness in handling data sparsity.

\noindent\textbf{IMDB-WIKI-DIR Benchmark.} IMDB-WIKI-DIR is a particularly challenging benchmark due to its dual challenges: label noise and data imbalance. Despite these intrinsic difficulties, our method demonstrates remarkable improvements over existing approaches as shown in Table~\ref{table:imdb}. Specifically, BSAM achieves the best overall performance with $6.811$ MAE on all samples, substantially outperforming conventional methods like RNC. The effectiveness of our approach is consistent across different data density regions: it achieves $6.294$ MAE for many-shot samples, and $10.823$ for medium-shot samples while maintaining robust performance for few-shot samples. The GM metric further demonstrates the superiority of our method, achieving a GM score of $3.765$, which is notably better than other methods. The results validate the effectiveness of our approach in handling complex age estimation tasks.

\noindent\textbf{NYUD2-DIR Benchmark.} We further evaluate our method on the NYUD2-DIR benchmark, with results shown in Table~\ref{table:NYU}. Our approach achieves state-of-the-art performance with the lowest RMSE of $1.272$ on All settings, outperforming strong baselines including Balanced MSE. Notably, our method demonstrates substantial improvements in the challenging Few-shot regime, indicating its effectiveness in handling the imbalance problem. For the $\delta_1$ metric, our method achieves competitive results of $0.727$ on All settings, demonstrating its capability to maintain high accuracy across different evaluation criteria.

\subsubsection{The bMAE metric}
As shown in Table~\ref{tab:bmae}, we present the results of the bMAE metric evaluation on both the AgeDB-DIR and IMDB-WIKI-DIR benchmarks. It is evident that BSAM consistently outperforms SQINV across different data density regions on both benchmarks in terms of bMAE measurement. Notably, in the few-shot region, where bMAE more effectively evaluates model performance, as stated in~\cite{ren2022balanced},  BSAM demonstrates significant improvements over SQINV.
\begin{table}[t]
\small
\centering
\caption{The bMAE metric~(lower is better) on AgeDB-DIR and IMDB-WIKI-DIR benchmark.}
\begin{tabular}{cccccc}
\specialrule{2\heavyrulewidth}{\abovetopsep}{\belowbottomsep}
   Datasets&Methods& All&Many&Med.&Few  \\ 
   \specialrule{1.5\heavyrulewidth}{\abovetopsep}{\belowbottomsep}
   \multirow{2}{*}{AgeDB}
   &SQINV & 7.075&5.955&7.203&9.278\\
   &BSAM & \textbf{6.734}& \textbf{5.801}&\textbf{6.340}&\textbf{8.890}\\\midrule
   \multirow{2}{*}{IMDB} 
   &SQINV & 11.838&6.673&13.718&29.826\\
   &BSAM &\textbf{11.282}&\textbf{6.356}&\textbf{11.333}&\textbf{26.466}\\ 
\specialrule{2\heavyrulewidth}{\abovetopsep}{\belowbottomsep}
    \end{tabular}
    \label{tab:bmae}
\end{table}

\begin{table}[t]
\small
\centering
\caption{The MAE of BSAM with vanilla regression loss on AgeDB-DIR benchmark.}
\begin{tabular}{ccccc}
\specialrule{2\heavyrulewidth}{\abovetopsep}{\belowbottomsep}
Methods      & All  &   Man. & Med.  &  Few   \\ 
\specialrule{1.5\heavyrulewidth}{\abovetopsep}{\belowbottomsep}
Vanilla &6.690  &5.959  & 7.740   &10.688  \\
Vanilla + Ours & \textbf{6.427} & \textbf{5.856} &  \textbf{7.116} & \textbf{9.915}   \\
        \specialrule{2\heavyrulewidth}{\abovetopsep}{\belowbottomsep}
    \end{tabular}
    \label{tab:vanilla}
\end{table}

\subsubsection{Combinations with different regression loss}
One of the distinctive features of BSAM is its flexible design regarding the choice of regression loss functions for parameter optimization. To verify this flexibility, we combined BSAM with vanilla regression loss ($l_1$), denoted as ``Vanilla + Our''.
Table~\ref{tab:vanilla} presents the comparative results on the AgeDB-DIR benchmark. As shown, integrating BSAM with the vanilla loss yields consistent improvements across all data density regions. Specifically, our approach reduces the overall MAE from $6.690$ to $6.427$.  These results indicate that BSAM's reweighting mechanism during perturbation calculation can enhance model performance independent of the specific form of loss function used, making it a versatile approach that can be combined with different regression losses to improve performance across different data density regions.

\subsubsection{Comparisons with different SAM}
To compare the effectiveness of our improvements to SAM, we evaluate various SAM-based methods as shown in Table~\ref{tab:diff_sam}. The standard SAM model improves the baseline MAE from $6.391$ to $6.230$, while the ImbSAM variant achieves an MAE of $6.184$. Our proposed method further reduces the MAE to $6.067$, demonstrating superior performance. While we observe performance variations across different data distribution regions, it is important to note that BSAM is designed to enhance generalization performance across the entire distribution. In contrast, ImbSAM primarily focuses on optimization for few-shot regions, which can potentially compromise overall performance across the complete dataset.
\begin{table}[t]
\small
\centering
\caption{The MAE of BSAM with different SAM  on AgeDB-DIR benchmark. }
\begin{tabular}{ccccc}
\specialrule{2\heavyrulewidth}{\abovetopsep}{\belowbottomsep}
Methods      & All  &   Man. & Med.  &  Few   \\ 
\specialrule{1.5\heavyrulewidth}{\abovetopsep}{\belowbottomsep}
SQINV        &6.391 & 5.955 &  7.155 & 8.390 \\
+ SAM~\cite{foret2021sharpnessaware} &  6.230  &\textbf{5.639} &7.313  &8.822 \\
+ imbSAM~\cite{zhou2023imbsam} &  6.184 &5.844 &6.799 &\textbf{7.695}  \\
+ BSAM &   \textbf{6.067} &5.801 & \textbf{6.304}  & 7.928 \\
\specialrule{2\heavyrulewidth}{\abovetopsep}{\belowbottomsep}
\end{tabular}
\label{tab:diff_sam} 
\end{table}

\begin{table}[t]
\small
\centering
\caption{The MAE of BSAM with different reweighting on AgeDB-DIR benchmark. }
\begin{tabular}{ccccc}
\specialrule{2\heavyrulewidth}{\abovetopsep}{\belowbottomsep}
Methods     & All  &   Man. & Med.  &  Few   \\ 
\specialrule{1.5\heavyrulewidth}{\abovetopsep}{\belowbottomsep}
SQINV        & 6.391 & 5.955 &  7.155 & 8.390 \\
\textit{INV}-BSAM & 6.146   & 5.806&6.668  &\textbf{7.921} \\
\textit{SQINV}-BSAM &   \textbf{6.067}  & \textbf{5.801} & \textbf{6.304}  & 7.928 \\
\specialrule{2\heavyrulewidth}{\abovetopsep}{\belowbottomsep}
\end{tabular}
\label{tab:diff_reweighting} 
\end{table}
\begin{table}[t]
\small
\centering
    \caption{Comparisons of $\lambda_{max}$ and $Tr(H)$ averaged over three experiments on the few-shot scenarios of AgeDB-DIR benchmark.}
    \label{tab:lambda}
    \begin{tabular}{ccc}
            \specialrule{2\heavyrulewidth}{\abovetopsep}{\belowbottomsep}
        Metric&SQINV&BSAM\\
\specialrule{1.5\heavyrulewidth}{\abovetopsep}{\belowbottomsep}
$\lambda_{max} \downarrow$&141.21&\textbf{86.51}\\
         $Tr(H) \downarrow$&653.53&\textbf{90.88} \\
\specialrule{2\heavyrulewidth}{\abovetopsep}{\belowbottomsep}
    \end{tabular}
\vspace{-5mm}
\end{table}

\subsubsection{Different Reweighting for the perturbation}
As shown in Table~\ref{tab:diff_reweighting}, we validate the impact of two approaches for calculating $w(k)$ on BSAM performance: inverse-frequency weighting~(\textit{INV}-BSAM) and square-root-inverse weighting~(\textit{SQINV}-BSAM). \textit{INV}-BSAM reduces the MAE to $6.146$, while \textit{SQINV}-BSAM further improves performance, achieving an MAE of $6.067$. Both methods show notable improvement compared to the baseline. This demonstrates that our approach is not dependent on a specific reweighting method but rather focuses on balancing the influence of data from all density regions during perturbation calculation.

\subsubsection{The effectiveness for low-density regions}
Table~\ref{tab:lambda} provides a quantitative analysis of the loss landscape geometry through two key metrics: the maximum eigenvalue $\lambda_{max}$ and the trace of the Hessian matrix $Tr(H)$. Lower values for both metrics indicate smoother, flatter loss landscapes associated with better generalization.
On the AgeDB-DIR benchmark in the few-shot scenarios, BSAM significantly outperforms SQINV. These dramatic improvements suggest that BSAM is particularly effective at creating flatter loss landscapes in low-density regions, which explains its superior generalization performance for underrepresented samples. 

\subsection{Ablation study}
\begin{figure}[t]
\centering
   \includegraphics[width=1.0\linewidth]{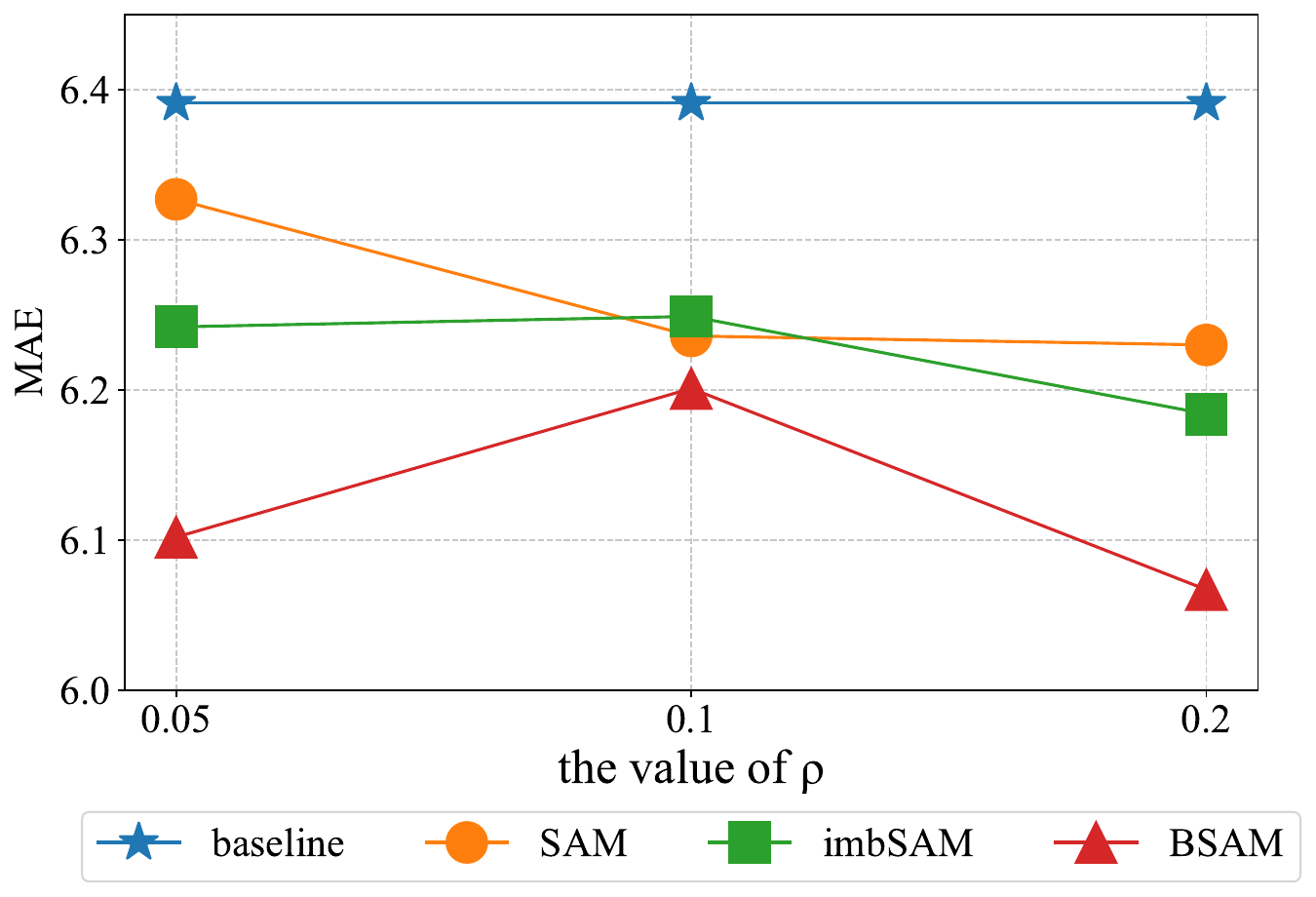}
   \caption{Ablation study for the value of $\rho \in \{0.05, 0.1, 0.2\}$ on AgeDB-DIR benchmark. }
\label{fig:ablation}
\vspace{-5mm}
\end{figure}

To evaluate the influence of the hyperparameter $\rho$, we conducted an ablation study on the AgeDB-DIR benchmark dataset. As shown in Figure~\ref{fig:ablation}, we compared the mean absolute error~(MAE) across different values of $\rho \in \{0.05, 0.1, 0.2\}$ for various methods: baseline, SAM~\cite{foret2021sharpnessaware}, imbSAM~\cite{zhou2023imbsam}, and our proposed BSAM approach. The results validate that our BSAM method effectively leverages sharpness to address imbalanced regression tasks, outperforming existing methods under different values of $\rho$.

\section{Conclusion}
\label{Conclusion}
In this paper, we have presented Balanced Sharpness-Aware Minimization~(BSAM), a novel approach that effectively addresses the challenge of imbalanced regression through principled integration of loss landscape sharpness and targeted reweighting mechanisms. Our analysis reveals the critical limitations of conventional SAM in handling underrepresented samples, leading to the development of a targeted reweighting mechanism that effectively balances model generalization across the entire data distribution. This simple yet effective approach maintains the computational efficiency of standard SAM while achieving superior performance across three challenging benchmarks including AgeDB-DIR, IMDB-WIKI-DIR, and NYUD2-DIR.

\noindent

\section{Acknowledgment}
This work is supported by the National Natural Science Foundation of China (No.62476051, No.62176047), the Sichuan Natural Science Foundation (No.2024NSFTD0041), and the Sichuan Science and Technology Program (No.2021YFS0374).

{
    \small
    \bibliographystyle{ieeenat_fullname}
    \bibliography{main}
}

\end{document}


\renewcommand{\thefigure}{S\arabic{figure}}
\renewcommand{\thetable}{S\arabic{table}}
\renewcommand{\thesection}{S\arabic{section}}

\maketitle


In this supplementary material, we provide comprehensive details to support our main manuscript. Specifically, we present: (1) detailed formulations of our evaluation metrics including MAE, GM, RMSE, and $\delta_1$, which thoroughly assess model performance across different aspects of the prediction distribution; (2) mathematical foundations of our reweighting strategies, including inverse-frequency weighting~(INV) and square-root-inverse weighting~(SQINV); and (3) the label distributions in the AgeDB-DIR, IMDB-WIKI-DIR and NYUD2-DIR benchmarks, demonstrating the prevalence and characteristics of regression imbalance in real-world vision tasks.
\section{Evaluation Metrics}
We employ multiple complementary metrics to evaluate the performance of our proposed method. Let $\mathcal{S} = \{(x_i, y_i)\}_{i=1}^N$ denote the test dataset where:
\begin{itemize}
\item $x_i \in \mathcal{X}$ represents the input image;
\item $y_i \in \mathcal{Y} \subset \mathbb{R}$ represents the ground truth regression target;
\item $\hat{y}_i \in \mathcal{Y}$ represents the predicted value;
\item $N$ denotes the total number of samples.
\end{itemize}
\subsection{Mean Absolute Error (MAE)}
MAE measures the average magnitude of errors in prediction without considering their direction:
\begin{equation}
\text{MAE} = \frac{1}{n}\sum_{i=1}^n |y_i - \hat{y}_i|.
\end{equation}
\subsection{Geometric Mean (GM)}
GM provides a measure of the central tendency of the absolute prediction errors by computing their geometric mean:
\begin{equation}
\mathrm{GM}=\left(\prod_{i=1}^N\left(\left|y_i-\hat{y}_i\right|\right)\right)^{\frac{1}{N}}.
\end{equation}
\subsection{Root Mean Square Error (RMSE)}
RMSE emphasizes larger errors due to its quadratic nature:
\begin{equation}
\text{RMSE} = \sqrt{\frac{1}{n}\sum_{i=1}^n (y_i - \hat{y}_i)^2}.
\end{equation}
where $(y_i - \hat{y}_i)^2$ represents the squared difference between the ground truth and predicted value.
\subsection{Threshold Accuracy ($\delta_1$)}
$\delta_1$ measures the percentage of predictions within a relative threshold:
\begin{equation}
\delta_1 = \frac{1}{n}\sum_{i=1}^n \mathbbm{1}[\max(\frac{y_i}{\hat{y}_i}, \frac{\hat{y}_i}{y_i}) < 1.25],
\end{equation}
where $\mathbbm{1}[\cdot]$ is the indicator function that returns 1 if the condition is true and 0 otherwise. And $1.25$ is the threshold for acceptable relative error.
\section{Reweighting Strategies}
Let $b_k$ denote the set of samples falling into the $k$-$th$ interval, and $n_k$ represents the number of samples in the interval $b_k$.

\subsection{Inverse-frequency Weighting (INV)}
INV assigns weights inversely proportional to the frequency of samples:
\begin{equation}
w_{\text{INV}}(k) = \frac{1}{n_k}.
\end{equation}
\subsection{Square-root-inverse Weighting (SQINV)}
SQINV provides a more moderate reweighting scheme:
\begin{equation}
w_{\text{SQINV}}(k) = \sqrt{\frac{1}{n_k}}.
\end{equation}
\section{Dataset Label Distribution Analysis}
\begin{figure}[t]
\centering
   \includegraphics[width=1.0\linewidth]{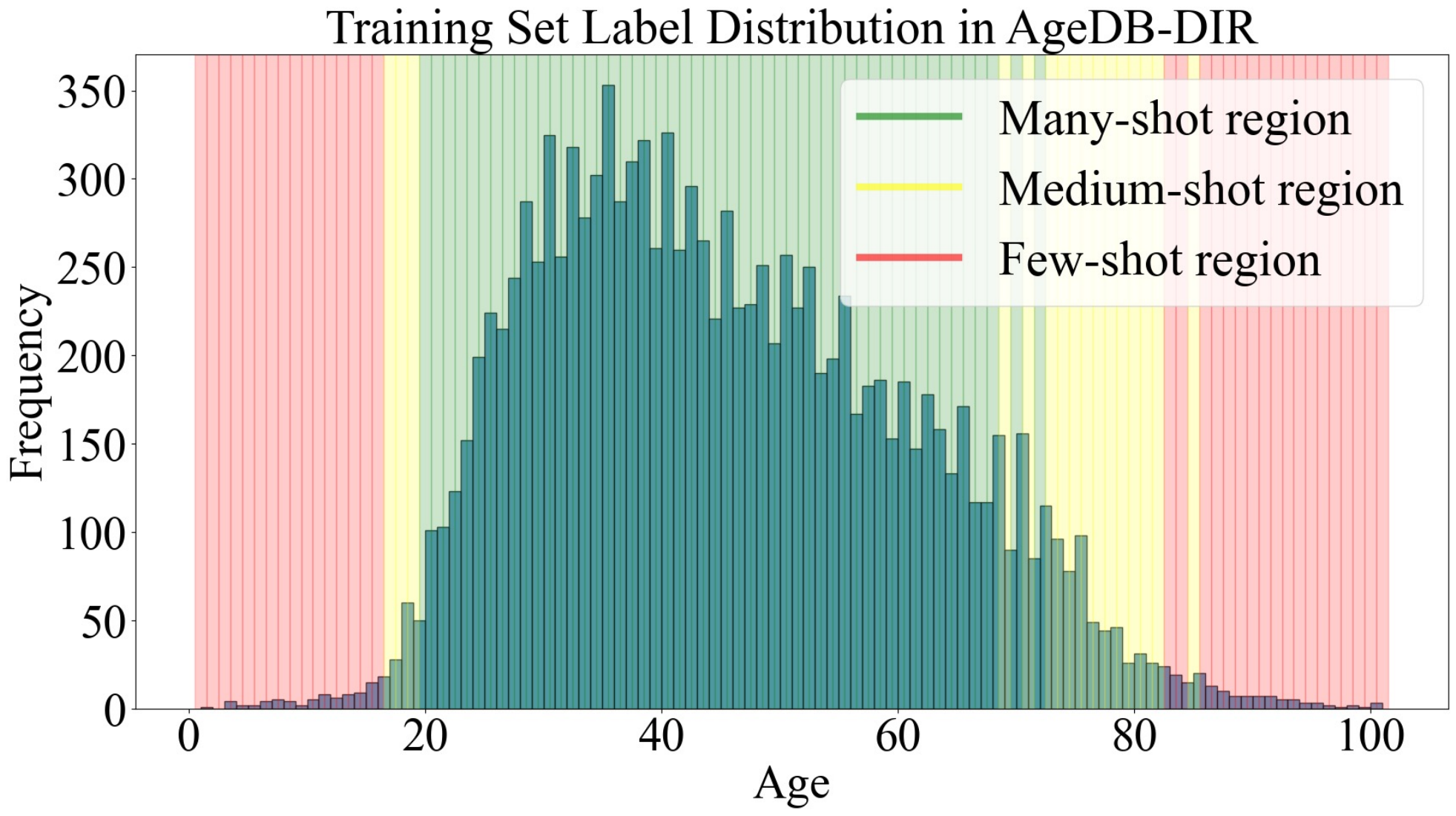}
   \caption{Training set label distribution in AgeDB-DIR. }
\label{fig:AgeDB}
\end{figure}

\begin{figure}[t]
\centering
   \includegraphics[width=1.0\linewidth]{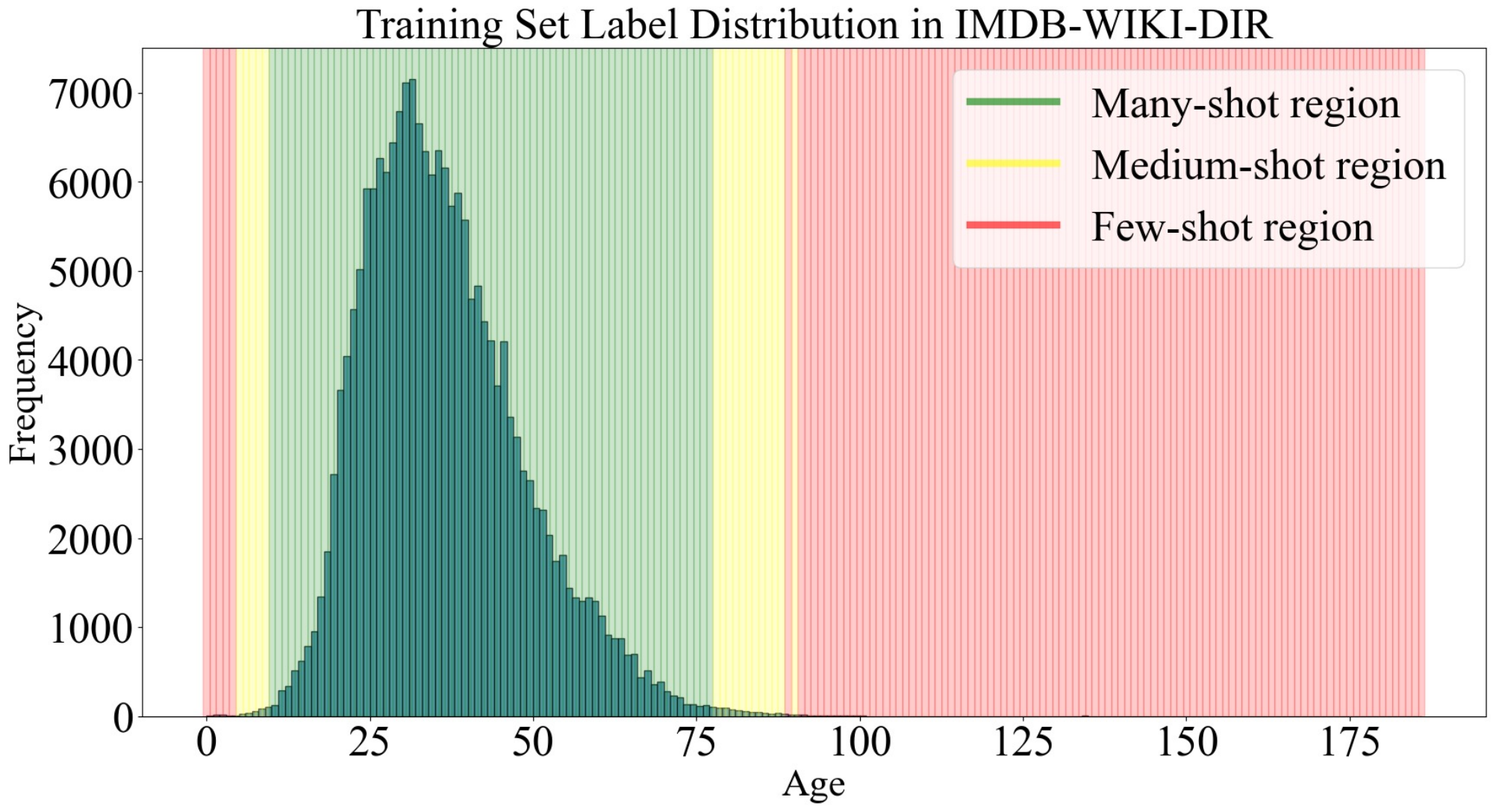}
   \caption{Training set label distribution in IMDB-WIKI-DIR. }
\label{fig:IMDB}
\end{figure}

\begin{figure}[t]
\centering
   \includegraphics[width=1.0\linewidth]{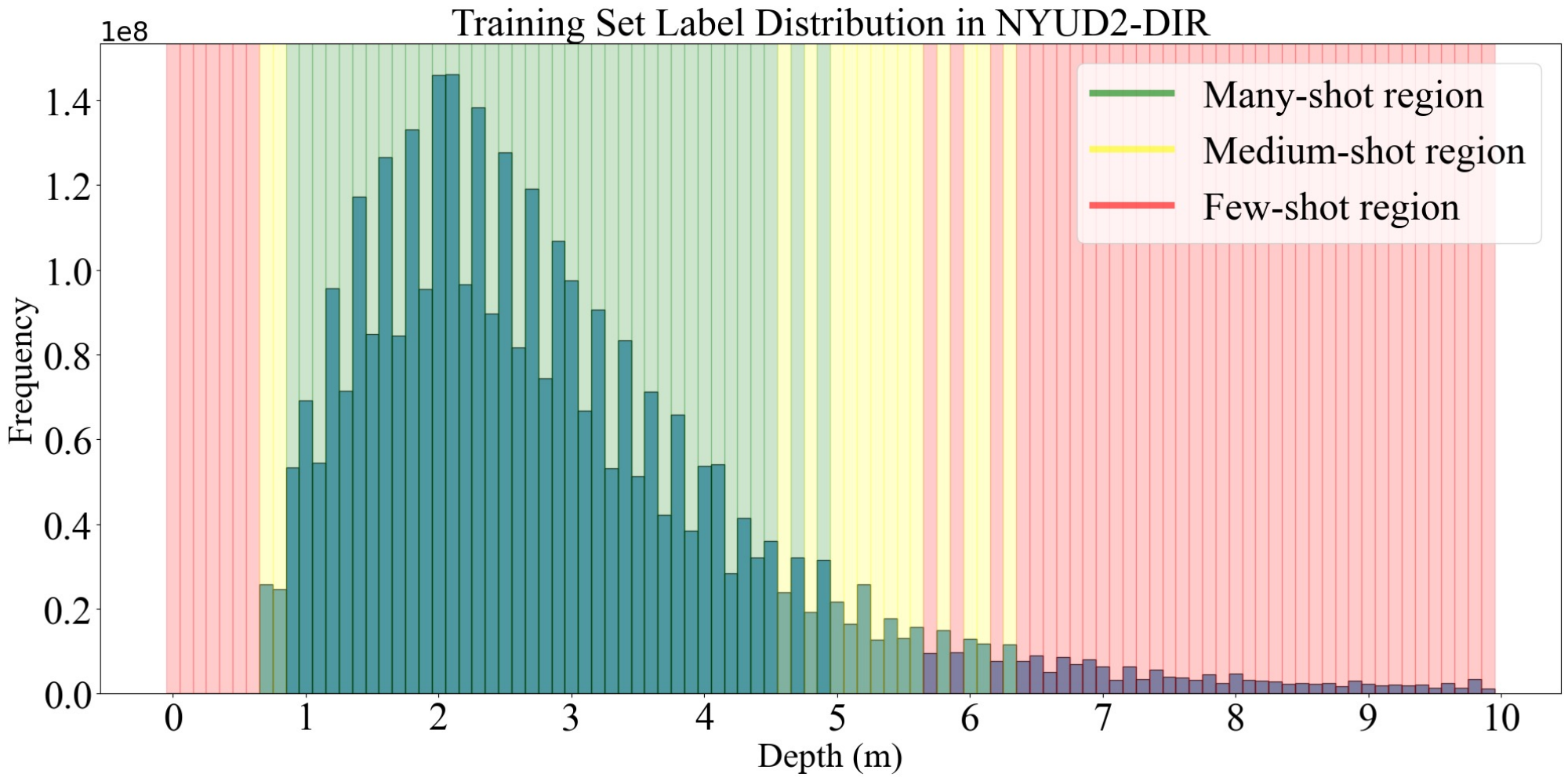}
   \caption{Training set label distribution in NYUD2-DIR. }
\label{fig:NYUD2}
\end{figure}

To comprehensively analyze the imbalanced regression problem in vision tasks, we investigate the label distributions of three representative datasets: AgeDB-DIR~(Figure~\ref{fig:AgeDB}), IMDB-WIKI-DIR~(Figure~\ref{fig:IMDB}) and NYUD2-DIR~(Figure~\ref{fig:NYUD2}).

\section{More Baselines}
We conducted additional experiments on the AgeDB dataset with more approaches, as shown in Tab.~\ref{tab:baseline}. The results further demonstrate the superiority of BSAM over stronger baselines.

\begin{table}[h]
\small
    \centering
    \caption{Comparisons with more baselines by MAE metric.}
    \begin{tabular}{c|cccc}
     Methods& All&Many&Med.&Few  \\ \hline
    Vanilla &6.690 &5.959&7.740&10.688 \\
   TERM~\cite{li2020tilted} & 6.518&5.935& 7.304& 9.848  \\
    RRT & 6.631&5.957&7.617&10.270 \\
    Focal-R~\cite{lin2017focal} & 6.565&5.837& 7.658 &10.427 \\
    BalanceMSE~(GAI)~\cite{ren2022balanced} & 6.541&6.036&6.927&10.243\\
    BalanceMSE~(BMC)~\cite{ren2022balanced} & 6.616&5.961&7.313&10.868\\
       \textbf{BSAM} & \textbf{6.067}& \textbf{5.801}&\textbf{6.304}&\textbf{7.928}\\ 
    \end{tabular}
    \label{tab:baseline}
\end{table}

\section{Limitation} 
Due to the limitations of existing imbalanced regression benchmarks, we have currently validated our method only on univariate imbalanced regression tasks. Multivariate imbalanced regression should be considered in future work. 

{
    \small
    \bibliographystyle{ieeenat_fullname}
    \bibliography{main}
}